\documentclass{article}
\usepackage{arxiv}
\usepackage[utf8]{inputenc} 
\usepackage[T1]{fontenc}    
\usepackage{hyperref}       
\usepackage{url}            
\usepackage{booktabs}       
\usepackage{amsfonts}       
\usepackage{nicefrac}       
\usepackage{microtype}      
\usepackage{amsmath}
\usepackage{cleveref}       
\usepackage{graphicx}
\usepackage[square,numbers]{natbib}
\usepackage{doi}
\usepackage{url}

\renewcommand{\headeright}{}
\renewcommand{\undertitle}{}

\begin{document}

\title{A General Model for Detecting Learner Engagement: Implementation and Evaluation}

\newcommand*\samethanks[1][\value{footnote}]{\footnotemark[#1]}
\author{
{\large Somayeh Malekshahi}\thanks{Department of Electrical and Computer Engineering, University of Tehran, Tehran, Iran}\\s.malekshahi@ut.ac.ir 
\and 
{\large \textbf{Javad M. Kheyridoost}}\\javadkheyridoost@gmail.com 
\and 
{\large \textbf{Omid Fatemi}}\samethanks[1]\\ofatemi@ut.ac.ir
}

\maketitle

\begin{abstract}
	Considering learner engagement has a mutual benefit for both learners and instructors. Instructors can help learners increase their attention, involvement, motivation, and interest. On the other hand, instructors can improve their instructional performance by evaluating the cumulative results of all learners and upgrading their training programs. This paper proposes a general, lightweight model for selecting and processing features to detect learners' engagement levels while preserving the sequential temporal relationship over time. During training and testing, we analyzed the videos from the publicly available DAiSEE dataset to capture the dynamic essence of learner engagement. We have also proposed an adaptation policy to find new labels that utilize the affective states of this dataset related to education, thereby improving the models' judgment. The suggested model achieves an accuracy of 68.57\% in a specific implementation and outperforms the studied state-of-the-art models detecting learners' engagement levels.
\end{abstract}


\section{Introduction}
Online learning facilitates access to asynchronous materials at any time and provides access to synchronous resources from anywhere. Consequently, the appropriate online learning materials to provide high interactivity and reliability are crucial \cite{Ref49}. The quality of the developing material in online learning directly impacts the effectiveness of the learning process \cite{Ref50}. Additionally, the learner's engagement with content and their interaction with the instructor and other learners during learning help to construct personal meaning  \cite{Ref51} for the learner who is at a distance from the instructor to promote learning and attain the desired learning outcomes. Analysis engagement of learners in an e-learning environment is the main concern in this work. We concentrate on automatically detecting learners' engagement levels to improve the quality of e-learning education. Awareness of learners’ engagement during online education provides a good insight into learning program performance and is a good criterion for instructors to level up the learning program based on the capabilities of learners.\\ 
In automatic learner engagement detection, models use video-based or frame-based evaluations. \cite{Ref14} used frame-by-frame labeling for recognition of student engagement by averaging dedicated labels to each frame as a final label of the 10-second video clip. In \cite{Ref15}, the level of student engagement was assessed based on the image, as the authors supposed that the engagement vary across 10-second video clips, making it difficult and sometimes inaccurate to assign a label to each video clip. In \cite{Ref16} although the videos of the classes have been used in groups, the dataset has been created based on photos. They also used emotional and behavioral aspects as engagement states. 
In DAiSEE dataset \cite{Ref17, Ref17_2}, the user's affective states were recognized, and benchmark results were presented using Static (Frame classification/prediction) and Dynamic (temporal video classification) models, both of which use images as inputs. In video classification, spatiotemporal relationships and actions between adjacent frames are recognized, and a label is assigned to the entire video.\\
In this paper, we use video-based evaluations, and our contributions are as follows: 1) We propose a general model for engagement recognition based on emotional expressions by preserving the sequential temporal relationship over time. 2) We apply different machine learning classifiers and find appropriate implementation settings on the DAiSEE dataset for competitive performance over state-of-the-art results on the same dataset 3) We analyze the DAiSEE dataset considering educational issues, propose an adaptive labeling for learner engagement, and train our model based on this adaptive labeling for better results.
Our paper is structured as follows:  First, we summarize the learner's engagement meaning and the previous works of automatic detection of learners' engagement levels. Subsequently, we present the main contribution of this paper. Next, we delve into the details of our experiments. Finally, we conclude and outline potential avenues for future research.

\section{Learner Engagement}
Educational psychologist Ralph Tyler's basic idea of the positive effects of time on task for learning was extended by many researchers like Pace (1980) as "quality of effort" for student success or Alexander Astin's theory of involvement (1984)\cite{Ref1}. George Kuh described engagement as a remarkable factor in student success as a set of elements that measure the amount of time and energy a student allocates to meaningful educational learning activities \cite{Ref1-1}. Groccia considered it as an indicator of student and institutional success and quality \cite{Ref2}. Additionally, there are other representations for student engagement related to their outcomes within the field of education.\\
Students’ engagement is beyond just attendance and involvement with learning activities. Interest, confidence, enjoyment, and a sense of belonging are signs of learning engagement \cite{Ref3,Ref4,Ref5}. Self-perceived is a notable point of engagement in learning distinguished by an instructor as a behavioral, cognitive, or emotional engagement \cite{Ref6,Ref7,Ref8}. Feeling and acting are both sides of measuring students’ engagement. Although using a range of measurable outcomes \cite{Ref9} is a criterion for engagement evaluation, learning a robust engagement measure is still challenging \cite{Ref10}.
Behavioral, emotional, and cognitive levels of student engagement are described in different dimensions, and considering these three categories in students' engagement process is complicated. The behavioral engagement regarding learner participation or effort needs to be continuously monitored. Detecting the affective or emotional level of learners' engagement is complex. Nevertheless, evaluating the level of interest, motivation improvement, and enjoyment or confusion and frustration provides insights into the degree of attention of learners during the instructor's presentation. Using such information, the instructors can encourage the learners not involved in the lesson at the right time. In this work, we use emotional facial expressions to detect emotional engagement. Engagement on a cognitive level is related to mental activity and needs evaluation metrics to assess \cite{Ref13}. 

\section{RELATED WORK}\label{RELATED WORK}
In \cite{Ref22}, a Deep Multi-Instance Learning (DMIL) framework was used to localize (in time) engaging/non-engaging parts in the online video lectures, which is useful for designing Massive Open Online Courses (MOOCs) video material. By assigning baseline scores based on the DMIL and computing localization as a MIL problem, they predicted students' engagement. Additionally, the authors created a new 'in the wild' database for engagement, consisting of  725,000 frames captured from 75 subjects of 102 videos, manually annotated by labelers for four levels of engagement. These are engagement intensity 0 as completely disengaged, engagement intensity 1 as barely engaged, engagement intensity 2 as engaged, and engagement intensity 3 as highly engaged. The duration of users' concentration was used as a criterion to determine the encouragement of the user to different parts of the video and the engagement level. \cite{Ref24} proposed a Deep Facial Spatiotemporal Network (DFSTN) to predict engagement, consisting of two parts. The first part extracted facial spatial features using the pre-trained SE-ResNet-50. In the second part, a more complicated architecture was used, employing Long Short Term Memory (LSTM) and attention mechanism to generate an attentional hidden state. The authors evaluated the performance of the proposed model using the DAiSEE dataset in four-class classification and obtained an accuracy of 58.84\%. They also tested the robustness of the model using the EmotiW-EP dataset.\\
\cite{Ref19} presented an end-to-end architecture using ResNet+TCN to detect students’ engagement levels. The hybrid network was trained in sequences of frames in videos of the DAiSEE dataset with a concentration on student engagement, with four levels of engagement: very low, low, high, and very high. They used an altered version of the network of \cite{Ref20} to extract the spatiotemporal features. The introduced model benefits from maintaining long-term history by TCN compared with LSTMs and GRUs. The imbalance problem in DAiSEE was resolved using a custom sampling strategy and a weighted cross-entropy loss function. In \cite{Ref25}, a CNN was introduced as an optimized lightweight model to detect students' learning engagement using facial expressions. This model was designed based on the ShuffleNet v2 architecture so that the authors utilized an attention mechanism to achieve the best performance. The ShuffleNet v2 0.5x structure was used to minimize computational complexity, different activation functions like the Leaky ReLU were examined to improve the accuracy of the model, and the attention module was embedded in the network to provide an optimized model. Finally, they obtained an accuracy of 63.9\% on the DAiSEE dataset with a focus on the state of engagement, with four levels of engagement as very low, low, high, and very high using the proposed model, Optimized ShuffleNet v2.\\
A new approach based on Neural Turing Machine for automatic engagement recognition was proposed in \cite{Ref10}. The approach used multi-modal feature combinations for learning the weights of features, which included eye gaze, facial action unit, head pose, and body pose features extracted from short videos using OpenFace 2.0 \cite{Ref40}. Experimental results demonstrated the accuracy of 61.3\% on the DAiSEE dataset in four levels of 0, 1, 2, and 3 from low to high on video engagement tags based on student participation. In \cite{Ref26}, authors proposed three models based on deep learning and used long-term memory in combination with EfficientNetB7 to detect students’ engagement levels in an online e-learning video. They evaluated their model’s accuracy using the self-developed dataset VRESEE and the public dataset DAiSEE with a concentration on student engagement, with four levels of engagement: very low, low, high, and very high. The spatial feature vectors related to the sequence of a video were represented with k-frames, but because of the difficulties of training, they found k=60 and k=40 as the best numbers of frames for DAiSEE and VRESEE datasets. They used three models for extracting temporal features: EfficientNet B7 with the Bidirectional LSTM, the hybrid EfficientNet B7 along with the Temporal Convolution Network (TCN), and the hybrid EfficientNet B7 and LSTM.\\
\cite{Ref27} explored the estimation of learning engagement in collaborative learning using gaze behavior and facial expression while the final student engagement was classified into four levels: high engagement, engagement, a little bit low engagement, and low engagement. Collaborative learning involves teaching activities that are handled by a group of participants, and interpersonal interaction between group members is a significant factor in the success of learning. To tackle facial analysis, the paper used a fuzzy logic of joint facial expression and gaze and introduced a multi-modal deep neural network (MDNN) to predict students' engagement. However, the lack of appropriate and publicly available datasets for collaborative learning environments poses a challenge in evaluating the results of such studies. In \cite{Ref28}, authors have a different view of measuring learner engagement. They did not consider the importance of the order of the behavioral and emotional states and just focused on the occurrence of these states. After extracting behavioral and influential features from video segments, they clustered feature vectors and decided based on the frequency of occurrence of the cluster centers. They achieved better accuracy on the IIITB Online SE dataset \cite{Ref29} and the DAiSEE student engagement dataset using the Bag of States (BoS) as a non-sequential approach. Among the four states of the annotated videos on the DAiSEE dataset, the focus was only on the engagement level classification in four levels of 0, 1, 2, and 3 as very low, low, high, and very high.\\ 
\cite{Ref30} used unlabeled classroom videos to detect the engagement level of the students. They utilized unsupervised domain adaptation for inferring student engagement and transferred features from an online setting to a classroom setting with unsupervised domain adaptation. They used Joint Adaptation Network and adversarial domain adaptation using Wasserstein distance and base models such as ResNet and I3D to resolve a binary classification problem. They achieved an accuracy of 68\% using the JAN network with RGB-I3D as the base model for the binary student engagement network. Also, an accuracy of 71\% was obtained using the adversarial domain adaptation with ResNet 50 as the feature extractor for predicting the engagement of the students. The DAiSEE dataset was a source dataset, and another was the target dataset used in the experiments. The two-level engagement detection approach considered only the affective state of engagement while ignoring other states. The label '0', the distracted label, was defined as the combination of very low and low engagement levels. The label '1', denoted as the engaged label, represented the combination of high and very high engagement levels.

\section{PROPOSED MODEL}
In this section, we first present our proposed model and a specific implementation for it, followed by a description of the various components of our learner engagement detection model. We introduce the available model for extracting emotional features, and then describe the dataset and the data preparation process to adapt it to our needs.

\subsection{General Model}
\begin{figure}[ht!]
\begin{center}
\includegraphics[width=10cm]{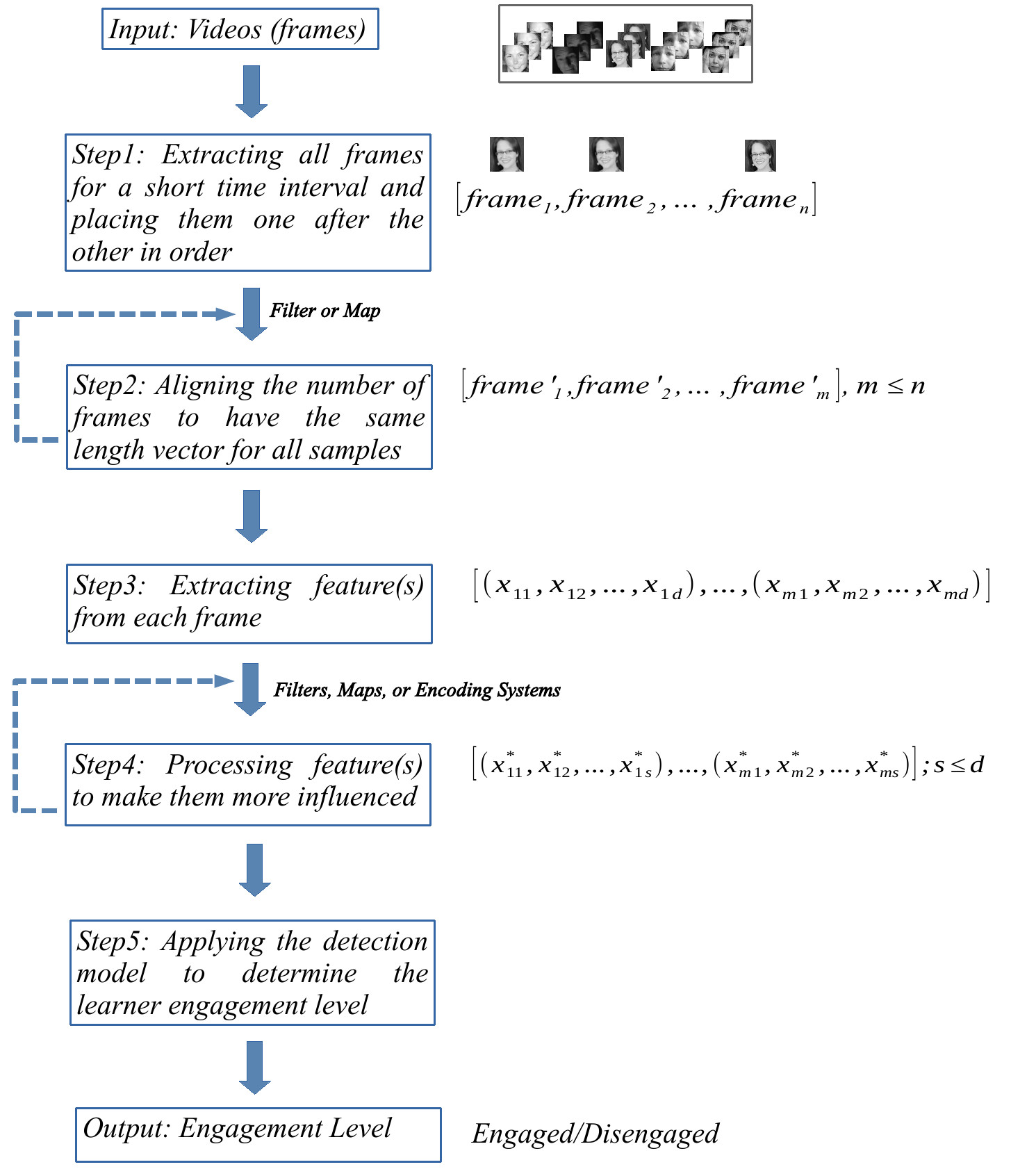}
\caption{Proposed general model for learner engagement detection. (The face images are adapted from \cite{Ref43,FER2013_Images})}
\label{fig:1}
\end{center}\vspace{-4mm}
\end{figure}

In this section, we introduce a general model for predicting learner engagement levels during a video clip, even in the presence of missing values. Missing features can result in biased estimations and negatively impact statistical results \cite{Ref44}, producing variable-length feature vectors. Here, we do not implement imputation or prediction solutions for missing features. Instead, as shown in Figure \ref{fig:1}, we propose a general model that processes the available features while preserving the sequential temporal relationship over time. All training and testing procedures are performed using the videos from the DAiSEE dataset. Within each 10-second video, we capture approximately 30 frames per second. However, the number of frames is not uniform across all samples, as we only evaluate frames that contain faces. Hence, different training input videos may have varying numbers of frames. We represent all frames of a video as a vector in Equation \ref{eq:1}, where ‘n’, the number of frames, is not uniform across all samples of the 10-second video clips. To address this issue, we apply filters or maps to shrink this vector to the same length for all samples and prepare them to use in classification models, resulting in a new vector represented in Equation \ref{eq:2}. An example of such a filter could be selecting a fixed number of frames within a specific time interval for all videos.

\begin{equation}\label{eq:1}
[frame_1, frame_2, ..., frame_n]
\end{equation}

\begin{equation}\label{eq:2}
[frame^{\prime}_1, frame^{\prime}_2, ..., frame^{\prime}_m] ; m \leq n
\end{equation}

Each frame in the vector is analyzed to represent a specific d-length feature vector using any model to extract these features like an emotional detection network, as expressed in Equation \ref{eq:3}. Features can be further manipulated by applying particular filters, maps, or encoding systems multiple times to extract the optimal s-length features for classification models, as shown in Equation \ref{eq:4}. The manipulation of features depends on their type and importance in the model, as determined by the dataset. Due to the role of the extracted features in the final diagnosis, a specific selection can be applied. Furthermore, the selected classification model in the learner engagement detection stage and the amount of data available for training the model play a role in establishing a balance between samples and features, thereby influencing the utilization of raw features.

\begin{equation}\label{eq:3}
[({x}_{11}, {x}_{12}, ..., {x}_{1 d}), ..., ({x}_{m 1}, {x}_{m 2}, ..., {x}_{md})]
\end{equation}

\begin{equation}\label{eq:4}
[({x}_{11}^{*}, {x}_{12}^{*}, ..., {x}_{1 s}^{*}), ..., ({x}_{m 1}^{*}, {x}_{m 2}^{*}, ..., {x}_{ms}^{*})] ; s \leq d
\end{equation}

The authors refer to filtering as a type of selection, mapping as a type of correspondence that is not necessarily reversible, and encoding as a type of transformation and conversion that can be decoded.

\subsection{Proposed Implementation} \label{ProposedImplementation}
\begin{figure}[ht!]
\begin{center}
\includegraphics[width=10cm]{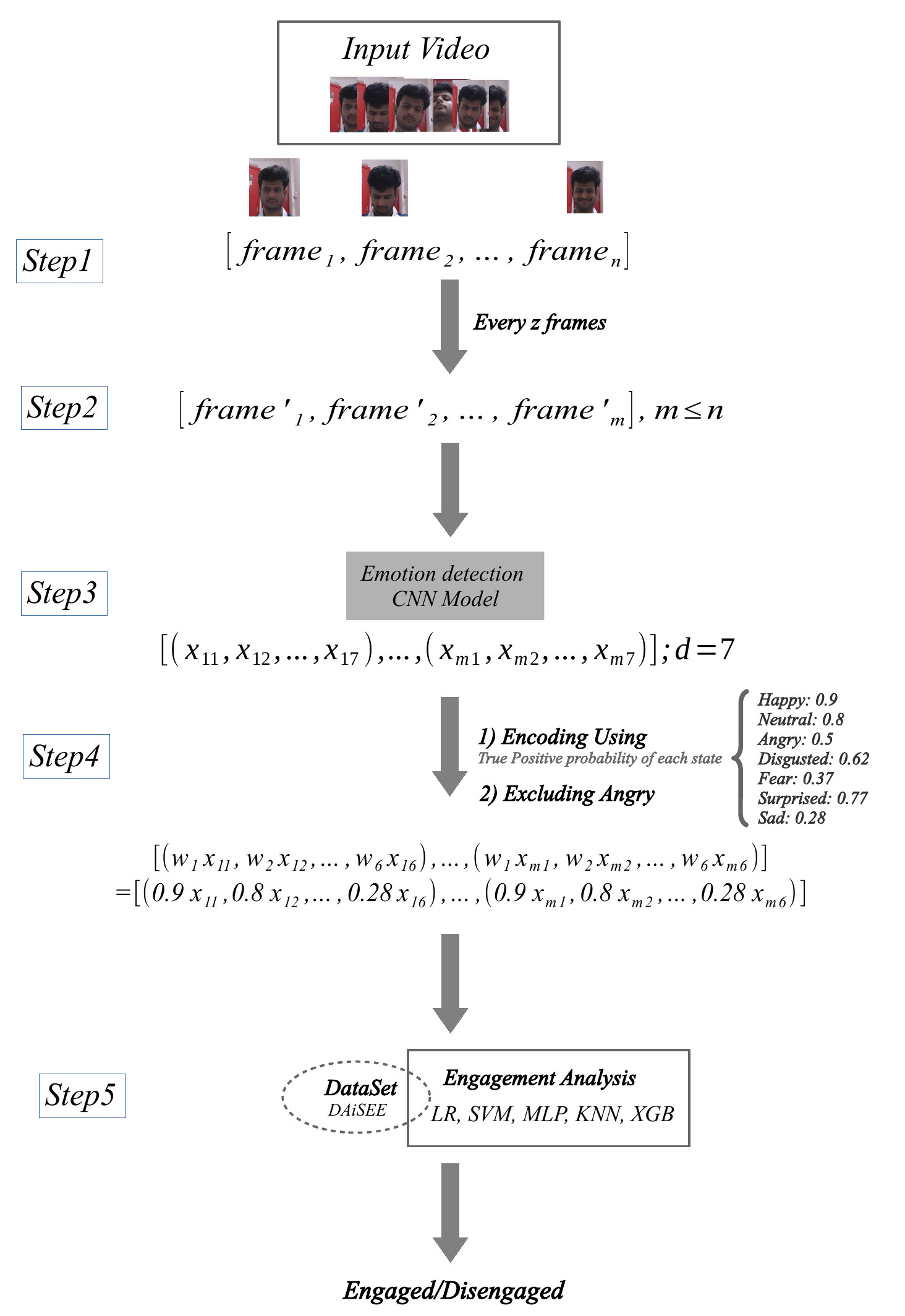}
\caption{Our specific implementation for the proposed model. (The face images are adapted from \cite{Ref17})}
\label{fig:2}
\end{center}\vspace{-4mm}
\end{figure}

In the following, we describe a particular implementation of the model, Figure \ref{fig:2}. We represent each frame of a video as a vector in Equation \ref{eq:1}, and then we reduce the number of frames in the samples to ensure that the feature vectors are of equal size. However, due to the importance of the ratio between the number of features and the number of samples in machine learning and the impact of this ratio on the performance and generalization ability of models, and due to the limited available data, we limit the number of frames as features for smooth training samples. Table \ref{tab:1} displays the number of frames in different videos, and we observe that if we consider more than 300 frames, we can only use 309 videos (each 10 seconds long), which is insufficient for training the classification models. Therefore, 'm' in Equation \ref{eq:2} cannot exceed 300.

\begin{table}[ht!]
  \caption{Extracted frames count of videos of the DAiSEE dataset}\vspace*{1ex}
  \label{tab:1}
  \begin{center}
  \begin{tabular}{| l | l | l | l | l | l | l |}
    \hline
    \multicolumn{1}{|c|}{\textbf{Frames}} & \multicolumn{1}{c|}{\textbf{{$>300$}}} & \multicolumn{1}{|c|}{\textbf{$\ge300$}} & \multicolumn{1}{|c|}{\textbf{{$\ge200$}}} & \multicolumn{1}{|c|}{\textbf{{$\ge100$}}} & \multicolumn{1}{|c|}{\textbf{{$\ge60$}}} & \multicolumn{1}{|c|}{\textbf{{$\ge40$}}} \\
    \hline
    \textbf{Videos}&309&6063&8079&8494&8647&8721\\
    \hline
  \end{tabular}
  \end{center}\vspace{-4mm}
\end{table}

To determine the appropriate number of frames per video, denoted as ‘m’ in Equation \ref{eq:2}, a filtering process is applied. Regarding the relatively low variations between neighboring frames, every ‘z’th frame is selected instead of all. Experiments conducted by \cite{Ref17} and ourselves have demonstrated that this approach does not significantly impact accuracy. However, we investigated different values for ‘z’ as a hyperparameter that requires searching to identify the optimal values during the training phase to customize the model for the specific dataset.\\
A seven-dimensional emotion vector is extracted from each frame using the emotion detection CNN described in Section \ref{EmotionDetectionModel}, Subsequently, a feature vector is constructed with a dimensionality of d=7, Equation \ref{eq:3}. We then apply an encoding process to each feature vector by multiplying each feature with the True Positive (TP) probability associated with each state recognized by the network, as explained in Section \ref{EmotionDetectionModel}. This approach gives more weight to the predictions of emotions with higher true positive rates and less weight to emotions with lower true positive rates. These values, denoted as $(w_1 , w_2, w_3, w_4, w_5, w_6, w_7)$ , correspond to the 'happy', 'neutral', 'surprised', 'disgusted', 'angry', 'fearful', and 'sad' states, and are (0.9, 0.8, 0.77, 0.62, 0.5, 0.37, 0.28). These coefficients prioritize the importance of each extracted emotion in the feature vector. While incorporating the knowledge of education science and psychology to choose the coefficients for prioritizing would make them more informed and influential, such considerations are beyond the scope of this paper.

\begin{equation}\label{eq:5}
\begin{split}
    [(x_{11}^{*}, x_{12}^{*}, ... , x_{17}^{*}) , ... , (x_{m1}^{*}, x_{m2}^{*}, ... , x_{m7}^{*}) ] \\
    = [(w_{1} x_{11} , w_{2} x_{12} , ... , w_{7} x_{17}) , ... , (w_{1} x_{m1} , w_{2} x_{m2} , ... , w_{7} x_{m7} )] \\
    = [(0.9 x_{11} , 0.8 x_{12} , ... , 0.28 x_{17}) , ... , ( 0.9 x_{m1} , 0.8 x_{m2} , ... , 0.28 x_{m7} )]
\end{split}
\end{equation}

Accurate detection of facial expressions relies on determining the neutral facial expression \cite{Ref52}. In human labeling, it is essential to consider neutral facial expressions of anger, anxiety, or happiness to avoid incorrect decisions. For instance, in videos of the DAiSEE dataset, some persons have a default frown, which is not a sign of anger, confusion, or frustration. Ignoring the neutral facial expression in a 10-second video can lead to incorrect recognition of emotions. While anger expression is correctly identified through feature-based learning in automatic expression detection, disregarding the person’s neutral facial expression results in inaccurate predictions in the engagement detection problem. Hence, we assess the impact of emotions on model estimation through feature importance analysis on the DAiSEE dataset. Each emotion is systematically removed as a feature, and the engagement prediction model is trained and evaluated both with and without each feature. The intensity of influence of these features indicates their contribution to the predictive capability of the classifier. Our experiments demonstrate that neglecting the default frown leads to issues while excluding the expression of anger improves prediction.\\
Using this extracted sequential information as features for each video, we try to preserve the dependency between frames' sequences in the learner engagement detection model. After feature selection in a specific condition, we need a model to be trained and then classify the learning engagement level. 
In this work, we use supervised learning classifiers to detect the learners' engagement level: Logistic Regression, Support vector machines (SVMs), Multi-Layer Perceptron (MLP), K-nearest neighbors (KNN), and eXtreme Gradient Boosting (XGBoost) are used. Logistic Regression is a statistical method for binary and multiclass classification. It is a Generalized Linear Model that attempts to use more data with almost the same effects to predict outcomes. It is a simple and efficient method used for classification problems. SVMs are supervised learning methods effective in high-dimensional spaces. SVMs depend on local data to find influential points and make more accurate decisions and are one of the most robust prediction methods. Multi-layer Perceptron (MLP) is a feed-forward neural network that can be used in classification problems that are not linearly separable. K-nearest neighbors (KNN) is a non-parametric, supervised machine learning algorithm we use for the classification problem. It is a simple and intuitive algorithm that labels a new sample based on its proximity to its k nearest neighbors in the feature space using distance metrics. eXtreme Gradient Boosting (XGBoost) is an effective and flexible machine learning algorithm based on the gradient boosting concept with an optimized implementation.

\subsection{Emotion Detection Model}\label{EmotionDetectionModel}
A CNN, designed in \cite{Ref41} with online implementation of \cite{Ref42}, is utilized to detect emotions. This network is trained on the FER-2013 dataset \cite{Ref43}, and includes seven emotional states: ‘angry’, ‘disgusted’, ‘fear’, ‘happy’, ‘sad’, ‘surprised’, and ‘neutral’. The highest true positive percentages for each respective emotion are as follows: 90\% for ‘happy’, 80\% for ‘neutral’, 77\% for ‘surprised’, 62\% for ‘disgusted’, 50\% for ‘angry’, 37\% for ‘fearful’, and 28\% for ‘sad’. The network receives an image as input that passes through a convolutional layer with a kernel size of 5 × 5 and a stride of 1. This layer is followed by a max-pooling layer of kernel size 3 × 3 with a stride of 2 in each dimension. A second convolutional layer with a kernel size of 5×5 and stride 1 feeds directly into the third convolutional layer with a kernel size of 4 × 4 and stride 1. Then, a dropout with a probability of 0.2 is applied to a fully connected linear layer. Finally, the softmax layer determines the most probable recognized emotion. We use this CNN for extracting emotional features from each frame by training it on the FER-2013 dataset for 100 epochs as \cite{Ref42} did to the accuracy convergence to the optimum. 

\subsection{Dataset} \label{Dataset}
DAiSEE (Dataset for Affective States in E-Environments) \cite{Ref17} is a video-level dataset annotated with affective states of 'Boredom', 'Engagement', 'Confusion', and 'Frustration', each with four levels of severity: Very Low, Low, High, and Very High. It contains 9068 video snippets with different illumination settings from 112 students aged 18-30, the Asian race, with 32 female and 80 male subjects. Authors in \cite{Ref17} evaluated the dataset with two types of temporal and static models based on CNNs and Top-1 accuracy was used to measure and analyze the results. According to the reported experimental results, 'Confusion' and 'Frustration' are more accurately detected, with 79.1\% accuracy for 'Frustration' using C3D FineTuning, 72.3\% accuracy for 'Confusion' using LRCN, 57.9\% accuracy for 'Engagement' using LRCN, and 53.7\% accuracy for 'Frustration' using LRCN.\\
While the dataset may not demonstrate satisfactory accuracy across all affective states, DAiSEE has been utilized in various studies related to learner engagement, as discussed in Section \ref{RELATED WORK}. It is widely employed because it is one of the few datasets publicly available with a video-level structure. The version of the dataset used in our study contains 2,455,390 frames from 8925 10-second videos, with the distribution of labels shown in Table \ref{tab:2}. The presented distribution exhibits slight variations compared to the reported results in \cite{Ref17},  stemming from using the dataset version accessed online \cite{Ref18}.

\begin{table}[ht!]
  \caption{Distribution of labels in the DAiSEE dataset }\vspace*{1ex}
  \label{tab:2}
  \begin{center}
  \begin{tabular}{| l | c | c | c | c |}
    \hline
    \multicolumn{1}{|c|}{\textbf{Affective State}} & \multicolumn{1}{c|}{\textbf{{Very Low}}} & \multicolumn{1}{|c|}{\textbf{Low}} & \multicolumn{1}{|c|}{\textbf{{High}}} & \multicolumn{1}{|c|}{\textbf{{Very High}}} \\
    \hline
    \textbf{Boredom}&3822&2850&1923&330\\
    \textbf{Confusion}&5951&2133&741&100\\
    \textbf{Engagement}&61&455&4422&3987\\
    \textbf{Frustration}&6887&1613&338&87\\
    \hline
  \end{tabular}
  \end{center}\vspace{-4mm}
\end{table}

DAiSEE is an imbalanced dataset, which means that the target class has an uneven distribution of observations. Furthermore, the distribution of this dataset is inconsistent and should be adapted to the context of the problem. Solutions to this issue can vary and depend on the context. For example, \cite{Ref30} changed the original four-level video engagement annotations to two classes. We also proposed an adaptation for a two-class classification problem, as shown in Table \ref{tab:3}. To this end, we will utilize a two-level labeling scheme, consisting of 'Engaged' and 'Disengaged' labels. 

\begin{table}[ht!]
  \caption{Adapting labels of the DAiSEE dataset to two-level engagement detection }\vspace*{1ex}
  \label{tab:3}
  \begin{center}
  \begin{tabular}{| l | l |}
    \hline
    \multicolumn{1}{|c|}{\textbf{Affective State}} & \multicolumn{1}{c|}{\begin{tabular}{@{}c@{}}\textbf{Policy} \end{tabular}}\\
    \hline
    \textbf{Engagement}&\begin{tabular}{@{}l@{}}$Engagement \leq 1 \rightarrow Disengaged$\\$Engagement \geq 2 \rightarrow Engaged$\end{tabular}\\
    \hline
    \textbf{Confusion}&\begin{tabular}{@{}l@{}}$Confusion == 0 \rightarrow Disengaged$\\$Confusion \geq 1 \rightarrow Engaged$\end{tabular}\\
    \hline
    \textbf{Frustration}&\begin{tabular}{@{}l@{}}$Frustration == 0 \rightarrow Disengaged$\\$Frustration \geq 1 \rightarrow Engaged$\end{tabular}\\
    \hline
    \begin{tabular}{@{}l@{}}\textbf{Engagement;}\\\textbf{Boredom}\end{tabular}&\begin{tabular}{@{}l@{}}$Engagement == 0 \rightarrow Disengaged$\\$Engagement == 3 \rightarrow Engaged$\\$Boredom \geq 2 \rightarrow Disengaged$\end{tabular}\\
    \hline
    \begin{tabular}{@{}l@{}}\textbf{Engagement;}\\\textbf{Confusion}\end{tabular}&\begin{tabular}{@{}l@{}}$Engagement == 0 \rightarrow Disengaged$\\$Engagement == 3 \rightarrow Engaged$\\$((Confusion == 0 \ or \ Confusion == 3) \ and $\\$\ \ \ \   Engagement == 1) \rightarrow Disengaged$\\$Otherwise \rightarrow Engaged $\end{tabular}\\
    \hline
    \begin{tabular}{@{}l@{}}\textbf{Engagement;}\\\textbf{Frustration}\end{tabular}&\begin{tabular}{@{}l@{}}$Engagement == 0 \rightarrow Disengaged$\\$Engagement == 3 \rightarrow Engaged$\\$((Frustration == 0 \ or \ Frustration == 3) \ and $\\$ \ \ \ \ Engagement == 1) \rightarrow Disengaged$\\$Otherwise \rightarrow Engaged$\end{tabular}\\
    \hline
    \begin{tabular}{@{}l@{}}\textbf{Confusion;}\\\textbf{Frustration}\end{tabular}&\begin{tabular}{@{}l@{}}$Frustration == 0 \ and \ Confusion == 0 \rightarrow Disengaged$\\$Frustration \geq 1 \ or \ Confusion \geq 1 \rightarrow Engaged$\end{tabular}\\
    \hline
    \begin{tabular}{@{}l@{}}\textbf{Confusion;}\\\textbf{Frustration}\\\textbf{Engagement}\end{tabular}&\begin{tabular}{@{}l@{}}$Engagement == 0 \rightarrow Disengaged$\\$Engagement == 3 \rightarrow Engaged$\\$((Frustration == 3 \ or \ Confusion == 3) \ and $\\$ \ \ \ \ Engagement == 1) \rightarrow Disengaged$\\$((Frustration == 0 \ or \ Confusion == 0) \ and $\\$ \ \ \ \ Engagement == 1) \rightarrow Disengaged$\\$Otherwise \rightarrow Engaged$\end{tabular}\\
    \hline
    \multicolumn{2}{|l|}
    {\begin{tabular}{@{}l@{}}\textbf{Severity level:} Very Low :0, Low: 1, High: 2, Very High: 3\\If a condition is not applicable, the label is assigned as 'None' (Not Determined)\end{tabular}}\\
    \hline
  \end{tabular}
  \end{center}\vspace{-4mm}
\end{table}

DAiSEE has the potential to be applied to various domains, such as e-learning, e-shopping, healthcare, autonomous vehicles, and even more fields \cite{Ref17}. Therefore, we will adapt the labeling scheme from this dataset to meet our specific requirements for education. Considering the importance of all affective states, namely 'Boredom', 'Engagement', 'Confusion', and 'Frustration' in education, and by drawing on our experience with various labeling schemes, we have found that meaningful combinations of different labels enhance the reliability of assigning a label to each video clip based on the two-level engagement detection. These labels are also relevant to learning engagement, making them a more appropriate choice for assessing our model's performance. 'Frustration' is a commonly experienced emotion during the process of learning, particularly in science and middle school environments. This emotion is a starting step toward encouraging students to become more engaged and learn more effectively. When teachers recognize signs of frustration, they can offer support to help students manage these negative emotions and move toward a more engaged state \cite{Ref33}. Additionally, confusion has been found to be a notable factor in the learning process in many previous studies \cite{Ref34,Ref35,Ref36,Ref37}. While frustration and confusion are both indicators of engagement, the severity of these negative emotions is crucial in analyzing the level of engagement. If confused students are not guided to manage these negative emotions, they may become frustrated or bored, leading to disengagement \cite{Ref38,Ref39}.\\ 
In this work, we consider all affective states and their severity in the DAiSEE dataset. Table \ref{tab:3} displays all label combinations considered. The first row indicates that, for the affective state of 'Engagement', we designate a severity level of Very Low and Low as label 0 (Disengaged), while High and Very High are assigned label 1 (Engaged). In 'Engagement; Confusion', fifth row, the severity level of Very Low for 'Engagement' is considered as Disengaged. Also, when 'Engagement' has a Very High value, it is labeled 1 (Engaged). In other cases, both 'Confusion' and 'Engagement' are in charge of determining the final label. The severity level of Very Low or Very High for 'Confusion' is set as Disengaged if the severity level of 'Engagement' is Very Low. Subsequent rows follow the same interpretation. If a condition is not applicable, the label is assigned as 'None' (Not Determined).\\
Applying our custom adapting policies, we calculated the distribution of labels based on the two-level engagement detection schema. The results in Table \ref{tab:4} reveal an imbalance for the different label combinations originating from the DAiSEE dataset, as shown in Table \ref{tab:2}. The most imbalanced case was ‘Engagement; Confusion’, with 8305 distinct labels, while the least imbalanced was ‘Confusion; Frustration’, with a difference of 1753 labels. Considering the significance of balanced data in classification, taking into account the label distribution results presented in Table \ref{tab:4} is crucial. Oversampling the minority class, under-sampling the majority class, focal loss, and weighted cross-entropy loss are accepted solutions to rectify the imbalance problem. In \cite{Ref19}, the authors addressed this issue using a weighted cross-entropy loss function, and we used random under-sampling based on the minimum number of samples labeled '0' and '1'. The last column of the Table \ref{tab:4} is the total number of balanced samples. For example, in the first row, there are 516 sample videos with the label '0', so we selected 516 samples randomly among videos with the label '1', and we had 1032 balanced video samples with the 'Engagement' label.

\begin{table}[ht!]
  \caption{Label distribution of the DAiSEE dataset adapted to two-level engagement detection labels}\vspace*{1ex}
  \label{tab:4}
  \begin{center}
  \begin{tabular}{| l | c | c | c | c|}
    \hline
    \multicolumn{1}{|c|}{\textbf{Affective State}} & \multicolumn{1}{c|}{\begin{tabular}{@{}c@{}}\textbf{Number of}\\\textbf{labels as ‘1’}\end{tabular}} & \multicolumn{1}{|c|}{\begin{tabular}{@{}c@{}}\textbf{Number of}\\\textbf{labels as ‘0’}\end{tabular}} & \multicolumn{1}{|c|}{\begin{tabular}{@{}c@{}}\textbf{Number of labels}\\\textbf{ as ‘Not determined’}\end{tabular}} & \multicolumn{1}{|c|}{\textbf{{total}}} \\
    \hline
    \textbf{Engagement}&8409&516&0&1032\\
    \hline
    \textbf{Confusion}&2974&5951&0&5948 \\
    \hline
    \textbf{Frustration}&2038&6887&0&4076\\
    \hline
    \begin{tabular}{@{}l@{}}\textbf{Engagement;}\\\textbf{Boredom}\end{tabular}&3987&1865&3073&3730\\
    \hline
    \begin{tabular}{@{}l@{}}\textbf{Engagement;}\\\textbf{Confusion}\end{tabular}&8615&310&0&620\\
    \hline
    \begin{tabular}{@{}l@{}}\textbf{Engagement;}\\\textbf{Frustration}\end{tabular}&8613&312&0&624\\
    \hline
    \begin{tabular}{@{}l@{}}\textbf{Confusion;}\\\textbf{Frustration}\end{tabular}&3586&5339&0&7172\\
    \hline
    \begin{tabular}{@{}l@{}}\textbf{Confusion;}\\\textbf{Frustration;}\\\textbf{Engagement}\end{tabular}&8535&390&0&780\\
    \hline    
  \end{tabular}
  \end{center}\vspace{-4mm}
\end{table}

After preprocessing and adapting the labels based on our labeling matching policy, and despite losing part of the data, we select balanced data based on the two-level labeling Engaged and Disengaged. The experimental results are explained in Section \ref{Results}. 

\section{EXPERIMENTAL RESULT}
\subsection{Performance Metric}

In designing classification models such as learners’ engagement level, evaluating the performance is a critical aspect. In supervised classification models, various evaluation metrics are available to measure how well the model will perform on unseen data. In this work, we use accuracy as an evaluation metric to compare our results with those obtained using the DAiSEE dataset.

\subsection{Results} \label{Results}

In this paper, we used supervised learning, wherein a model is trained to make predictions or classifications based on labeled training data. In supervised learning, the training dataset comprises input samples (features) and their corresponding output labels, which serve as ground truth or supervision for the learning process. The goal is to acquire a mapping function that can effectively generalize from the training data, enabling accurate predictions or classifications on unseen or future data instances.\\
We evaluated our results in two ways. First, we used the default 'Engagement' affective state of DAiSEE to evaluate our model, following the approach used in other papers. Second, as we detailed the structure and challenges of the DAiSEE dataset in Section \ref{Dataset}, we assigned a label for each video based on our adapted two-level labeling approach. We then preprocessed and prepared the data for training the engagement detection models. To reproduce experimental conditions, we used the origin partitioning of the dataset. We combined all videos of train and validation sets for training and used the test videos exclusively for model evaluation. We balanced them by selecting an equal number of the videos labeled as engaged and those labeled as disengaged. In addition, to have more normally distributed data and a relatively similar scale, we used standardization. As described in Section \ref{ProposedImplementation}, we investigated hyperparameter, the number of skipped frames, ‘z’. Among the different values, we achieved an accuracy of 68.57\% using every 30th frames (z=30), with 60 features. During the 10-second videos, temporal variations are generally minimal. This suggests that selecting different segments is likely to yield similar outcomes, aligning with our expectations. \\
Given that prior works on comparisons of engagement detection models used the 'Engagement' label from the DAiSEE dataset, in our experiments of the 10 frames and 60 features tested, KNN achieved the highest accuracy of 67.64\%, followed by XGB with 63.73\%, MLP, SVM, and LR with the accuracies of 62.73\%, 61.82\% and, 60.91\%, shown in Table \ref{tab:5}. Based on our proposed adaptation methods for a two-class classification problem described in Section \ref{Dataset} and presented in Table \ref{tab:3}, the results are depicted in Table \ref{tab:6}. Among the combinations of effective states of 'Engagement', 'Confusion', and 'Frustration' that are more relevant to learning engagement, discussed in detail in Section \ref{Dataset},  the combination of 'Engagement' and 'Confusion' is more accurate and outperforms the individual states. The classification report of the best result of the 'Engagement; Confusion' with the accuracy of 68.57\% using KNN is shown in Table \ref{tab:7}.

\begin{table}[ht!]
  \caption{The accuracy of learner engagement detection for balanced data on the DAiSEE dataset for 'Engagement'}\vspace*{1ex}
  \label{tab:5}
  \begin{center}
  \begin{tabular}{| l | c | c | c | c | c |}
    \hline
    \multicolumn{1}{|c|}{\textbf{Affective State}} & \multicolumn{1}{c|}{\textbf{KNN}} & \multicolumn{1}{|c|}{\textbf{XGB}} & \multicolumn{1}{|c|}{\textbf{MLP}} & \multicolumn{1}{|c|}{\textbf{SVM}} & \multicolumn{1}{|c|}{\textbf{LR}}\\
    \hline
    \textbf{Engagement}&\textbf{67.64}\%&63.73\%&62.73\%&61.82\%&60.91\%\\
    \hline 
  \end{tabular}
  \end{center}\vspace{-4mm}
\end{table}

\begin{table}[ht!]
  \caption{The accuracy of learner engagement detection for balanced data on the DAiSEE dataset for different adapting policies of selecting affective states}\vspace*{1ex}
  \label{tab:6}
  \begin{center}
  \begin{tabular}{| l | c |}
    \hline
    \multicolumn{1}{|c|}{\textbf{Affective State}} & \multicolumn{1}{c|}{\textbf{KNN}} \\
    \hline
    \textbf{Engagement}&67.64\%\\
    \hline
    \textbf{Confusion}&55.45\%\\
    \hline
    \textbf{Frustration}&56.99\%\\
    \hline
    \begin{tabular}{@{}l@{}}\textbf{Engagement;}\\\textbf{Boredom}\end{tabular}&53.10\%\\
    \hline
    \begin{tabular}{@{}l@{}}\textbf{Engagement;}\\\textbf{Confusion}\end{tabular}&\textbf{68.57}\%\\
    \hline
    \begin{tabular}{@{}l@{}}\textbf{Engagement;}\\\textbf{Frustration}\end{tabular}&66.13\%\\
    \hline
    \begin{tabular}{@{}l@{}}\textbf{Confusion;}\\\textbf{Frustration}\end{tabular}&56.71\%\\
    \hline
    \begin{tabular}{@{}l@{}}\textbf{Confusion;}\\\textbf{Frustration;}\\\textbf{Engagement}\end{tabular}&65.81\%\\
    \hline    
  \end{tabular}
  \end{center}\vspace{-4mm}
\end{table}

\begin{table}[ht!]
  \caption{The classification report of 'Engagement; Confusion', and KNN of test videos of the balanced data}\vspace*{1ex}
  \label{tab:7}
  \begin{center}
  \begin{tabular}{| c | c | c | c |}
    \hline
    \multicolumn{1}{|c|}{\textbf{Label}} & \multicolumn{1}{c|}{\textbf{{precision}}} & \multicolumn{1}{|c|}{\textbf{recall}} & \multicolumn{1}{|c|}{\textbf{{f1-score}}} \\
    \hline
    \textbf{Disengaged}&0.67&0.74&0.70\\
    \textbf{Engaged}&0.71&0.63&0.67\\
    \hline
  \end{tabular}
  \end{center}\vspace{-4mm}
\end{table}

\subsection{Comparison} \label{Comparison}
In this section, we compare the performance of our method with the related works using more complicated models to classify the learner’s engagement level, Table \ref{tab:8}. All results were evaluated by accuracy as the performance metric using the DAiSEE dataset. \cite{Ref26} introduced three different models based on EfficientNet B7 and LSTM, Bi-LSTM, and TCN and obtained 67.48\%, 66.39\%, and  64.67\% for accuracy metric, respectively. In \cite{Ref25} accuracy of 63.9\% was reported for the proposed optimized lightweight model based on ShuffleNet v2 architecture. \cite{Ref19} achieved 61.15\% and 63.9\% for proposed models of combining ResNet as a spatial feature extractor with LSTM and TCN to understand the changes in different frames, respectively. The result of using the pre-trained SE-ResNet-50, LSTM, and attention mechanism as Deep Facial Spatiotemporal Network, is an accuracy of 58.84\%, \cite{Ref24}. The accuracy of 61.3\% is the achievement of \cite{Ref10} utilized the Neural Turing Machine and extracted multi-modal features for automating engagement recognition. In our models the accuracy of 68.57\% is with 'Engagement;Confusion' state and 67.64\% is with ‘Engagement’ state both with KNN on balanced data. 

\begin{table}[ht!]
  \caption{Results of different models on videos of the DAiSEE dataset}\vspace*{1ex}
  \label{tab:8}
  \begin{center}
  \begin{tabular}{| c | c | c |}
    \hline
    \multicolumn{1}{|c|}{\textbf{Methods}} & \multicolumn{1}{c|}{\textbf{Accuracy}}&\multicolumn{1}{c|}{\textbf{Reference}}\\
    \hline
    {KNN and CNN}&68.57\%&Proposed model\textsuperscript{*}\\
    \hline
    {KNN and CNN}&67.64\%&Proposed model\textsuperscript{**}\\
    \hline
    {EfficientNet B7 and LSTM}&67.48\%&\cite{Ref26}\\
    \hline
    {Bi-LSTM, and TCN}&66.39\%&\cite{Ref26}\\
    \hline
    {ShuffleNet v2}&63.9\%&\cite{Ref25}\\
    \hline
    {ResNet and TCN}&63.9\%&\cite{Ref19}\\
    \hline
    {ResNet and LSTM}&61.15\%&\cite{Ref19}\\
    \hline
    {Neural Turing Machine}&61.3\%&\cite{Ref10}\\
    \hline
    \begin{tabular}{@{}c@{}}{SE-ResNet50 + LSTM  with}\\{Global Attention}\end{tabular}&58.84\%&\cite{Ref24}\\
    \hline
    \multicolumn{3}{|l|}
    {\begin{tabular}{@{}l@{}}\textsuperscript{*}based on the adapting policy of ‘Engagement; Confusion’\\\textsuperscript{**}based on the default 'Engagement' affective state \end{tabular}}\\
    \hline 
  \end{tabular}
  \end{center}\vspace{-4mm}
\end{table}

\section{CONCLUSION}

This paper proposes a general model for detecting learners' engagement with a custom feature selection method. The processing of frames to select influential features from the variable-length feature vectors, using a video-based dataset while preserving temporal dependencies, resulted in marginally higher accuracy than existing approaches, surpassing them by approximately one percent. While the improvement may seem modest, it is noteworthy given the incremental gains often observed in such systems and demonstrates the effectiveness of our model in achieving competitive performance with a more streamlined approach. Also, incorporating the affective states related to education and creating new labels to adapt videos to educational contexts has helped achieve more reliable results. Moreover, in the learner engagement classification part, we streamline the training process significantly by selecting statistical and efficient models, compared with neural networks and deep learning models that require a significant amount of labeled data and computational resources. The simplicity of this model makes it applicable for use in synchronous classes as an online learner engagement detector. For future work, we would like to explore alternative methods to address missing features to obtain smooth feature vectors by considering all learners to develop a model that accurately evaluates the level of learner engagement in synchronous classes. Additionally, we intend to explore datasets containing longer clips to provide a closer representation of real-world educational classrooms.

%\bibliographystyle{IEEEtran} 
%\bibliography{references}
% Generated by IEEEtran.bst, version: 1.14 

\end{document}